\documentclass[11pt]{article}
\usepackage{paper}

\usepackage{amsmath, amsfonts, amssymb}
\usepackage{float}

\usepackage{adjustbox}
\newlength\commaheight

\usepackage{tikz}

\title{Understanding Robust Machine Learning for Nonparametric Regression with Heavy-Tailed Noise}
\begin{document}

\author[1]{Yunlong Feng}
\author[2]{Qiang Wu}
\affil[1]{Department of Mathematics and Statistics, State University of New York at Albany}
\affil[2]{Department of Mathematics, University of Tennessee, Knoxville}

\date{}

\maketitle

\begin{abstract}
\noindent We investigate robust nonparametric regression in the presence of heavy-tailed noise, where the hypothesis class may contain unbounded functions and robustness is ensured via a robust loss function $\ell_\sigma$. Using Huber regression as a close-up example within Tikhonov-regularized risk minimization in reproducing kernel Hilbert spaces (RKHS), we address two central challenges: (i) the breakdown of standard concentration tools under weak moment assumptions, and (ii) the analytical difficulties introduced by unbounded hypothesis spaces. Our first message is conceptual: conventional generalization-error bounds for robust losses do not faithfully capture out-of-sample performance. We argue that learnability should instead be quantified through prediction error, namely the $L_2$-distance to the truth $f^\star$, which is $\sigma$-independent and directly reflects the target of robust estimation. To make this workable under unboundedness, we introduce a \emph{probabilistic effective hypothesis space} that confines the estimator with high probability and enables a meaningful bias--variance decomposition under weak $(1+\epsilon)$-moment conditions. Technically, we establish new comparison theorems linking the excess robust risk to the $L_2$ prediction error up to a residual of order $\mathcal{O}(\sigma^{-2\epsilon})$, clarifying the robustness--bias trade-off induced by the scale parameter $\sigma$. Building on this, we derive explicit finite-sample error bounds and convergence rates for Huber regression in RKHS that hold without uniform boundedness and under heavy-tailed noise.  Our study delivers principled tuning rules, extends beyond Huber to other robust losses, and highlights prediction error, not excess generalization risk, as the fundamental lens for analyzing robust learning. Moreover, it provides a pathway toward understanding heavy-tailed kernel methods and deep models within a unified theory of robustness.  
\end{abstract}

\section{Introduction and Overview}

In this paper, we study robust machine learning for nonparametric regression problems, particularly those involving heavy-tailed noise. The goal of nonparametric regression learning is to learn a function in a nonparametric manner from a hypothesis space, typically through risk minimization. In the presence of heavy-tailed noise, robust risk minimization is often employed to enhance the stability and resilience of the learned model.  

Robust regression has been extensively explored in parametric statistics, usually under linear assumptions on the regression function and specific noise distributions; see, e.g., \cite{huber1973robust,holland1977robust,rousseeuw1984least,yohai1987high}. The field of robust statistics has developed as a dedicated discipline to address such challenges \cite{huber2009robust,maronna2006robust,hampel2011robust,rousseeuw2005robust}. However, in nonparametric contexts where these assumptions no longer hold, theoretical understanding of robust learning remains limited, despite its widespread application in modern data science. Building on our prior work \cite{feng2020learning,feng2022statistical1,feng2020framework,feng2025tikhonov} and inspired by the work in \cite{hu2013learning,fan2016consistency}, this study seeks to unify and advance the understanding by exemplifying a well-established risk minimization approach, namely nonparametric Huber regression, as a case study to draw broader insights into robust methods. We begin by revisiting the formulation of robust regression learning via risk minimization.  

\subsection{Robust learning for regression via risk minimization}
Let $\mathcal{X}\subset\mathbb{R}^d$ be an input space, $\mathcal{Y}=\mathbb{R}$ the output space, and $\mathbf{z}=\{(x_i,y_i)\}_{i=1}^n$ i.i.d.~observations drawn from an unknown distribution $\rho$ over $\mathcal{X}\times\mathcal{Y}$. We consider the classic regression model \cite{vapnik1998statistical},
\begin{align}\label{data_generating}
y = f^\star(x) + \varepsilon,    
\end{align}
where, without loss of generality, it is assumed that $\mathbb{E}[\varepsilon|x]=0.$

Risk minimization approaches regression by selecting from a hypothesis space $\mathcal{H}$ the candidate function that minimizes the empirical risk  
\[
\sum_{i=1}^n \ell(y_i - f(x_i)),
\]  
where $\ell: \mathbb{R}\to\mathbb{R}_+$ is a loss function measuring the pointwise goodness-of-fit of predicting $y_i$ with $f(x_i)$. The hypothesis space $\mathcal{H}$ may consist of candidate functions such as subsets of a reproducing kernel Hilbert space (RKHS) used in kernel methods, or functions represented by neural networks. To balance flexibility in learning with protection against overfitting, a common approach is to incorporate regularization. That is, one selects the best candidate by minimizing a penalized empirical risk, e.g.,  
\[
\sum_{i=1}^n \ell(y_i - f(x_i)) + \lambda \|f\|^2,
\]  
with $\lambda>0$ a regularization parameter and $\|f\|$ an appropriate norm of $f$.  

For regression under heavy-tailed noise, robustification is typically achieved by replacing the loss function $\ell$ with a robust loss $\ell_\sigma$ that incorporates a scale parameter $\sigma$ serving as a tuning knob during training \cite{huber2009robust,andrews2015robust}. Such losses are Lipschitz continuous, either bounded or at most linearly growing, and may be convex or nonconvex. Popular examples include Huber \cite{huber1964robust}, Tukey \cite{tukey1960survey}, Welsch \cite{dennis1978techniques}, Geman--McClure \cite{geman1987statistical}, and Barron losses \cite{barron2019general}, many of which originate from classical robust statistics.  

Among them, the Huber loss is perhaps the most widely used in machine learning due to its convexity and favorable optimization properties. Accordingly, to make the exposition concrete, we focus on risk minimization with the Huber loss, using a reproducing kernel Hilbert space $\mathcal{H}_K$ induced by a reproducing kernel $K$ as the hypothesis space. It should be emphasized, however, that our analysis and results extend broadly to a wide class of robust risk minimization strategies across different hypothesis spaces.

\subsection{A close-up example: nonparametric Huber regression}

By means of regularization, risk minimization implements Huber regression in $\mathcal{H}_K$ via the scheme 
\begin{align}\label{empirical_target_function}
f_{\mathbf{z}} \;=\; \arg\min_{f\in\mathcal{H}_K} \Bigg\{ \frac{1}{n}\sum_{i=1}^n \ell_\sigma\big(y_i - f(x_i)\big) + \lambda \|f\|_K^2 \Bigg\},
\end{align}  
where $\lambda>0$ is a regularization parameter, $\|\cdot\|_K$ denotes the RKHS norm, and $\ell_\sigma$ is the Huber loss defined by  
\begin{align*}
\ell_\sigma(t) =   
\begin{cases}
2\sigma|t|-\sigma^2, & |t|\geq \sigma, \\
t^2, & \text{otherwise}.
\end{cases}
\end{align*}

Denote by $f_{\sigma,\lambda}$ the population counterpart of $f_\mathbf{z}$, given by  
\begin{align}\label{population_version}
f_{\sigma,\lambda} \;=\; \arg\min_{f\in\mathcal{H}_K} \Big\{ \mathbb{E}\,\ell_\sigma(y-f(x)) + \lambda\|f\|_K^2 \Big\}.
\end{align}
For notational simplification, we define the empirical and population risks, respectively, as  
\begin{align*}
\mathcal{R}_\mathbf{z}(f) \;=\; \frac{1}{n}\sum_{i=1}^n \ell_\sigma\big(y_i-f(x_i)\big), 
\qquad 
\mathcal{R}(f) \;=\; \mathbb{E}\,\ell_\sigma(y-f(x)).
\end{align*}
The out-of-sample generalization ability of a learning machine $f$ is typically quantified by bounding either the excess risk $\mathcal{R}(f)-\mathcal{R}(f_\mathcal{M})$, where $f_\mathcal{M}$ minimizes the population risk over all measurable functions $\mathcal{X}\to\mathbb{R}$, or $\mathcal{R}(f)-\mathcal{R}(f^\star)$, where $f^\star(x)=\mathbb{E}[y|x]$. In least squares regression, when the noise $y-f^\star(x)$ has finite variance, one has $f^\star=f_\mathcal{M}$ almost surely \cite{cucker2007learning}. This equivalence, however, does not hold for general risk minimization schemes induced by $\ell_\sigma$.  

The question of out-of-sample learnability is central to machine learning and will also be a primary focus of this study. On the other hand, unlike in classical robust statistics where $\sigma$ is treated as a fixed scale parameter, here it plays the role of a tuning parameter. Accordingly, we will investigate how $\sigma$ influences the learnability of $f_\mathbf{z}$ when the hypothesis space functions are not uniformly bounded, as in \eqref{empirical_target_function}. These two directions form the main threads of the present work.

\subsection{Understanding robust learning: knowns and unknowns}

Robust regression has long been recognized as an effective tool for mitigating the influence of outliers and heavy-tailed noise \cite{huber2009robust}. In classical settings, where the hypothesis space is uniformly bounded, there exists a well-developed theory ensuring consistency, favorable bias--variance trade-offs, and meaningful generalization error bounds \cite{cucker2007learning,steinwart2008support}. In such cases, robust losses such as the Huber loss yield stable and efficient estimators, particularly when the data are light-tailed \cite{fan2024noise}.  In addition, optimization and statistics are well aligned in this regime: the learning objective is well behaved, the empirical process can be controlled with standard tools, and the resulting bounds translate cleanly into out-of-sample guarantees.

In contrast, many practical machine learning problems arise in nonparametric regimes where the hypothesis space is infinite-dimensional and functions are not uniformly bounded. This setting introduces substantial technical challenges. Standard concentration inequalities and risk analyses often fail when candidate functions can take arbitrarily large values, because envelope arguments no longer apply and rare but extreme observations can dominate empirical averages \cite{cucker2007learning,steinwart2008support}. Heavy-tailed outcomes pose a second difficulty. When only weak moments are assumed, fluctuations of empirical risks can be highly uneven and classical comparisons between training objectives and prediction performance may be unreliable. While weak moment conditions, such as Assumption~\ref{moment_assumption} introduced in Section~\ref{sec:moment}, are widely used to model realistic noise, they introduce an additional source of bias that must be explicitly accounted for in the analysis. Balancing this bias against the variance reduction delivered by robust losses becomes a central theme in the heavy-tailed setting. A further, and more subtle, issue concerns the alignment between the optimization target implied by a robust loss and the statistical target of inference. In general, the population minimizer of a robust objective does not coincide with the underlying truth, even when the hypothesis space is very rich. Consequently, decreases in robust risk do not necessarily translate into improvements in prediction accuracy. This mismatch suggests that prediction error, rather than excess robust risk, should be the primary yardstick for learnability in robust nonparametric regression. It also underscores the importance of comparison principles that connect robust risk to prediction error while making explicit the residual gap due to robustification. The role of tuning is also different from the classical picture. The scale parameter is often adjusted during training rather than fixed in advance, and it interacts with regularization and model complexity in nontrivial ways. Smaller scales increase robustness to outliers but can enlarge population bias relative to the statistical target; larger scales reduce this bias but weaken down-weighting of extreme residuals. Regularization, in turn, controls the effective capacity of the fitted function class and indirectly influences the range of fitted values that drive the behavior of robust losses. It was unclear   how to treat these knobs jointly and coordinate them to achieve fast rates. Finally, while the Huber loss is a natural and widely used choice because of its convexity and favorable optimization properties, robust learning in practice often employs other losses with different tail behaviors. Understanding which qualitative properties, such as linear growth in the tails and quadratic behavior near zero, are essential for learnability helps delineate when results carry over beyond a single loss and when they do not.

\subsection{Our contributions to the literature}
Our contributions to the literature can be summarized as follows: 

First, we introduce a \emph{probabilistic effective hypothesis space} \(\mathcal{H}_\sigma\) that confines the estimator in a probabilistic sense. This localization restores concentration without assuming uniform boundedness, places the empirical and population target functions inside the \emph{same} effective set, and enables a new bias–variance decomposition tailored to robust losses. The construction explicitly reflects the interplay among the scale parameter, regularization, and tail heaviness, and it functions as the technical backbone for subsequent analysis.

Second, we establish comparison theorems that link excess robust risk to the \(L_2\) prediction error up to an explicit residual of order \(\mathcal{O}(\sigma^{-2\epsilon})\). The results are two‑sided, uniform over the effective class, and make precise the robustness–bias trade‑off induced by the scale parameter. In particular, they yield \emph{asymptotic mean calibration} within \(\mathcal{H}_\sigma\): decreases in robust risk translate into prediction improvements once the residual is controlled by a suitable (typically diverging) choice of the scale. This clarifies when robustification helps and how far one can push it before bias dominates.

Third, we derive nonasymptotic error bounds and convergence rates for Huber regression under weak moment assumptions and without uniform boundedness of the hypothesis space. The guarantees expose how tail heaviness, approximation properties of the RKHS, and localized capacity interact, and they deliver concrete tuning advice by coupling the scale and regularization parameters. In particular, our rates are accompanied by explicit choices of the scale parameter as a function of sample size and  the regularization parameter, illustrating when and why robust regression outperforms least squares in nonparametric regimes.

Finally, we offer a conceptual reframing of learnability for robust methods: \emph{prediction error}, rather than excess robust risk, should be the primary metric. We formalize this by separating approximation effects, driven by the hypothesis space, regularization, and the marginal distribution of inputs, from a distributional robust‑bias term controlled by the scale parameter in the loss. Although our analysis is conducted on Huber regression for concreteness, the underlying ingredients of our analysis, localization via a probabilistic effective hypothesis space, uniform comparison between robust risk and prediction error, and heavy‑tail–aware concentration, depend only on qualitative properties of the loss, not on convexity. Consequently, the framework extends to general robust risk minimization schemes, including nonconvex redescending losses such as Tukey, Welsch, and Geman–McClure, as well as pseudo‑Huber and Barron‑type losses. Therefore, our study provides a template for analyzing robust supervised machine learning schemes, including  kernel methods and overparameterized models.

\noindent\textbf{Road-map of the paper}: Section~2 sets up robust regression via risk minimization, highlights the obstacles posed by heavy-tailed noise and non‑uniformly bounded hypothesis spaces, and presents the nonparametric Huber scheme in an RKHS together with its basic properties. Section~3 develops the core theory: we introduce the probabilistic effective hypothesis space that localizes the analysis, and we prove comparison theorems that relate excess Huber risk to \(L_2\) prediction error, clarifying when reductions in robust risk imply predictive gains. Section~4 provides the finite‑sample analysis, deriving explicit error bounds and convergence rates under weak moment conditions and translating them into joint tuning guidance for the scale and regularization parameters. Section~5 broadens the perspective with discussions on implications, extensions to other robust losses, and practical considerations in heavy‑tailed regimes. Section~6 concludes with a summary of the main insights.

\noindent\textbf{Notation and convention}. Throughout this paper, it is assumed that $\|f^\star\|_\infty\leq M$ with $M>0$ a constant. We write  
\(u \preccurlyeq v\) if there exists a universal constant 
\(C>0\) such that $u  \le Cv$. For $R>0$, we write $\mathcal{B}_R := \{ f\in\mathcal{H}_K : \|f\|_K \le R \}$.
We use the embedding $\|f\|_\infty \le \kappa \|f\|_K$ with $\kappa:=\sup_{x\in\mathcal X} \sqrt{K(x,x)}<\infty$.

\section{Knowing What is Being Learned is Essential in Robust Learning}\label{sec::what_learned}
The empirical target function $f_\mathbf{z}$ learned from the robust risk minimization \eqref{empirical_target_function} depends not only on the regularization parameter $\lambda$ but also on the tuning parameter $\sigma$ from the loss function $\ell_\sigma$. Even with a sufficiently large hypothesis space, its population version, say $f_{\sigma,\lambda}$ or $f_\mathcal{M}$, however, may not be the same as the truth function $f^\star$ that one seeks. This raises a basic question: \emph{what quantity should we track to assess the out-of-sample learnability of $f_\mathbf{z}$}?

Our first message is that analysis should be anchored to $f^\star$, not to $f_{\sigma,\lambda}$ or $f_\mathcal{M}$. Conventional excess robust risk,
\[
\mathcal{R}(f_\mathbf{z})-\mathcal{R}(f_\mathcal{M}) \quad \text{or} \quad
\mathcal{R}(f_\mathbf{z})-\mathcal{R}(f_{\sigma,\lambda}),
\]
is generally \emph{insufficient} for measuring learnability because the target of the robust objective is $\sigma$-dependent and can be biased away from $f^\star$.  This can be illustrated using the example presented in \cite{feng2022statistical1}, which considers the regression model
	\begin{align*}
	y=f^\star(x)+\varepsilon,
	\end{align*}
	where $\varepsilon$ is a zero-mean noise variable with the probability density function
	\begin{align*}
	p_{\varepsilon}(t)=
	\begin{cases}
	\frac{1}{2}e^{-(t+\frac{1}{4})}, & \hbox{ if } t\geq -\frac{1}{4},\\[1ex]
	e^{2(t+\frac{1}{4})}, & \hbox{ if } t<-\frac{1}{4}.
	\end{cases}
	\end{align*}
Consider a simplified scenario in which the hypothesis space is a bounded subset of a certain $\mathcal{H}_K$ and $\lambda=0$. As shown in \cite{feng2022statistical1}, in this case, even with a fixed $\sigma$ value, $f_\sigma(x)\neq f^\star(x)$ for all $x\in\mathcal{X}$. Furthermore, the convergence of $\mathcal{R}(f_\mathbf{z})$ to $\mathcal{R}(f_{\sigma,\lambda})$ does not imply the convergence of $f_\mathbf{z}$ to $f^\star$, even in a probabilistic sense. As detailed later, this occurs because introducing $\sigma$ into the loss function leads to the dependence of the empirical target function  $f_\mathbf{z}$ and the target function $f_{\sigma,\lambda}$ on $\sigma$, which further introduces inherent biases to the two functions and makes the target function $f_{\sigma,\lambda}$ intractable when the value of $\sigma$ varies. This implies that $\mathcal{R}(f_\mathbf{z}) - \mathcal{R}(f_{\sigma,\lambda})$ may not be suitable for quantifying the out-of-sample learnability of robust learning models. One might then attempt to measure $\mathcal{R}(f_\mathbf{z})-\mathcal{R}(f^\star)$ directly. This, however, is also problematic because the functional $\mathcal{R}(\cdot)$ itself depends on $\sigma$. As $\sigma$ changes, the scale of the objective and the notion of optimality change with it, so the same numerical gap can correspond to very different predictive accuracies. Put differently, using a $\sigma$-dependent risk to assess proximity to a $\sigma$-independent target conflates the effect of tuning with out-of-sample performance.

For these reasons, we adopt the \emph{$L_2$ prediction error} associated with $f^\star$ as the primary yardstick:
\[
\|f_\mathbf{z}-f^\star\|_{2,\rho}^2
=\int_{\mathcal{X}} \big(f_\mathbf{z}(x)-f^\star(x)\big)^2\,\mathrm{d}\rho_x(x).
\]
This metric is invariant to the choice of robust loss, directly reflects the true target $f^\star$, and, as we show later, can still be tightly related to robust risk through comparison inequalities that hold in an appropriate effective class.

To summarize, robust learning requires keeping track of \emph{what} is being learned. Excess robust risk relative to a loss-dependent population minimizer can be misleading, especially when $\sigma$ is tuned. Prediction error with respect to $f^\star$ provides a $\sigma$-independent and semantically correct criterion.

\section{Are We Afraid of Unbounded Hypothesis Spaces or Heavy-Tailed Conditional Distributions in Robust Learning?}
Regularized robust learning models  involve unbounded hypothesis spaces, which comprise functions that are not uniformly bounded. In particular, the non-uniform boundedness of functions in the hypothesis space, coupled with the unboundedness and heavy-tailedness of the response variable, poses significant challenges in quantifying the learnability of robust learning models and undermines the applicability of existing machine learning theories. In this section, we examine these perceived challenges, providing insights into their underlying causes, and outline our proposed strategies and framework for advancing the understanding of robust machine learning models.

\subsection{Identifying key barriers in the analysis}
Generally speaking, existing studies in machine learning theory focus primarily on unregularized and regularized convex risk minimization schemes, with an emphasis on their out-of-sample prediction ability typically quantified through generalization error or prediction error. To be precise and to facilitate our discussion that follows, we start from the following general unregularized risk minimization scheme
\begin{align*}
f_\mathbf{z}=\arg\min_{f\in\mathcal{H}}\frac{1}{n}\sum_{i=1}^n\ell(y_i-f(x_i)),  
\end{align*}
where $\ell$ represents a convex loss function and $\mathcal{H}$ is a pre-determined hypothesis space. In this case, the convergence of the excess generalization error to zero can be guaranteed under light-tailed assumptions on the distribution of $y|x$, mild conditions on $\mathcal{H}$ and the loss function $\ell$ \cite{cucker2007learning,steinwart2008support}.

Robust regression changes this picture in two intertwined ways. First, in nonparametric regimes the hypothesis space is often infinite-dimensional and functions are not uniformly bounded. Without a uniform envelope, standard symmetrization and Bernstein-type arguments can break down, and bias–variance decompositions that assume boundedness become uninformative. As a consequence, traditional excess-risk bounds may no longer reflect predictive performance, and it becomes preferable to work with prediction error as the main criterion. Second, heavy-tailed noise invalidates many light-tailed heuristics. Under weak moment assumptions, empirical risks may be dominated by rare but extreme observations, slowing concentration and complicating uniform comparisons between training objectives and prediction error. Robust losses mitigate variance inflation by down-weighting extremes, but they also introduce an additional bias. Balancing this additional bias against the variance reduction delivered by robustification is a central difficulty in analysis.

\subsection{Towards the heavy-tailed: Modeling the conditional distributions using weak moment conditions}\label{sec:moment}
In robust regression, classical tail assumptions, such as finite variance or sub-Gaussian behavior, can be unnecessarily strong. A more flexible and realistic alternative is to impose only a weak moment requirement on the conditional distribution of the response. The following assumption formalizes this idea.

\begin{assumption}\label{moment_assumption}
There exists an absolute constant $\epsilon>0$ such that the conditional $(1+\epsilon)$-moment of $y$ given $x$ is square integrable with respect to $\rho_x$; that is,
\[
a(x)\;=\;\mathbb{E}\!\left[\,|y|^{\,1+\epsilon}\mid x\right]\in L^2_{\rho_x}.
\]
\end{assumption}

Assumption~\ref{moment_assumption} is deliberately mild and has been widely adopted in robust estimation; e.g., \cite{catoni2012challenging,audibert2011robust,brownlees2015empirical,lugosi2019mean,fan2024noise}. It does not require a full second moment, nor any form of exponential tail decay. Instead, it asks that a slightly higher-than-first conditional moment exists and that the function $x\mapsto a(x)$ be square integrable on the covariate space. This brings several advantages. First, it accommodates heavy-tailed settings, including conditional Pareto- or $t$-like behaviors, as long as the tail index exceeds $1+\epsilon$ on average across $x$. Second, it subsumes familiar light-tailed regimes: when noise is sub-Gaussian or sub-exponential, the assumption holds for arbitrarily large~$\epsilon$.

This weak moment condition allows us to quantify the parts of the risk contributed by large residuals without assuming uniform boundedness of the hypothesis class or finite conditional variances. It is weak enough to cover heavy-tailed phenomena encountered in practice, yet strong enough to enable the tail bounds and concentration estimates that underpin the comparison theorems and finite-sample guarantees developed in later sections.

\subsection{Towards the heavy-tailed: Avoiding confronting the conditional distributions in bias analysis}

As stated in Section \ref{sec::what_learned}, in the analysis of the out-of-sample generalization in robust learning, the prediction error, quantified using a $\sigma$-independent distance metric between $f_{\mathbf{z}}$ and $f^\star$, is preferred over the excess generalization error. This preference is particularly relevant when studying the impact of $\sigma$ on learnability. As discussed there, the $L_2$-distance serves this purpose effectively. Employing the $L_2$-distance as a metric to assess the predictive performance of $f_{\mathbf{z}}$ offers a multitude of advantages. On the other hand, in machine learning, error analysis of risk minimization estimators typically involves examining bias, arising from the choice of hypothesis space, the underlying data distribution, and the regularity of the target function, and variance, which is caused by randomized sampling and the capability of the hypothesis space.

In this subsection, we highlight another benefit of using the $L_2$-distance that is especially remarkable in the context of robust learning: it simplifies bias analysis by circumventing the need to directly deal with the conditional distribution of $y|x$, which is likely to be heavy-tailed in robust learning problems.

To proceed with our discussion, recall that the sample-free version of $f_{\mathbf{z}}$, denoted as $f_{\sigma,\lambda}$ can be equivalently written as follows:
\begin{align}\label{population_version_huber_regularized}
f_{\sigma,\lambda}:=\arg\min_{f\in\mathcal{H}_K} \mathcal{R}(f)-\mathcal{R}(f^\star)+\lambda \|f\|_K^2,   
\end{align}
and
\begin{align}
\mathcal{D}(\lambda,\sigma):= \mathcal{R}(f_{\sigma,\lambda})-\mathcal{R}(f^\star)+\lambda \|f_{\sigma,\lambda}\|_K^2.  
\end{align}
Here, $\mathcal{D}(\lambda,\sigma)$ represents the bias of $f_{\mathbf{z}}$ with conceptually infinite samples. Clearly, it depends not only on the tuple $(\mathcal{H}_K, f^\star, \rho_x)$ but also on the conditional distribution $\rho_{y|x}$ and the scale parameter $\sigma$. We further introduce a reference function in $\mathcal{H}_K$ that aligns with $f_{\sigma,\lambda}$ in terms of the choice of $\lambda$: 
\begin{align}\label{population_version_ls_regularized}
f_\lambda:=\arg\min_{f\in\mathcal{H}_K} \|f-f^\star\|_{2,\rho}^2+\lambda \|f\|_K^2
\end{align}
and correspondingly, denote 
\begin{align}
\mathcal{D}(\lambda):=  \|f_\lambda-f^\star\|_{2,\rho}^2+\lambda \|f_\lambda\|_K^2,  
\end{align}
which represents the approximation capability of the tuple $(\mathcal{H}_K, \lambda, \rho_x)$ towards $f^\star$.

The following theorem tells us that with the help of the reference function $f_\lambda$, the estimation of the approximation error $\mathcal{D}(\lambda,\sigma)$ can be boiled down to the estimation of $\mathcal{D}(\lambda)$, which is independent of the conditional distribution $\rho_{y|x}$, and an additional bias term that solely depends on $\sigma$. In other words, the introduction of the reference function $f_\lambda$ in the bias analysis allows us to strip the dependence on $\rho_{y|x}$ and $\sigma$, pertaining only to the tuple $(\mathcal{H}_K, \lambda, \rho_x)$.

\begin{theorem}\label{thm:right}
Let Assumption \ref{moment_assumption} hold. Under the restriction  $\sigma \geq \max(1, M)$,
the following inequality holds:
\begin{align}\label{eq:rightonly}
\mathcal{D}(\lambda,\sigma)\leq \mathcal{D}(\lambda) + C' \sigma^{-2\epsilon},
\end{align}
where \(C'\) is an absolute positive constant independent of \(f\) or \(\sigma\).
\end{theorem}

\subsection{Towards the unbounded: Introducing probabilistic effective hypothesis spaces}
As stated above, another essential barrier in analyzing and understanding the robust estimator $f_{\mathbf{z}}$ lies in the non-uniform boundedness of functions in the hypothesis space $\mathcal{H}_K$. In fact, it is easy to see that one can roughly upper bound $\|f_{\mathbf{z}}\|_\infty$ in the following way:     

\begin{lemma}\label{rough_bound}
Let $f_\mathbf{z}$ be produced by \eqref{empirical_target_function}. Then, it holds that    $\|f_\mathbf{z}\|_\infty\leq \frac{2\kappa^2\sigma}{\lambda}$.   
\end{lemma} 
\begin{proof}
From the risk minimization scheme \eqref{empirical_target_function} and the Lipschitz continuity property of $\ell_\sigma$, we have
\begin{align*}
\lambda\|f_\mathbf{z}\|_K^2\leq \frac{1}{n}\sum_{i=1}^n \ell_\sigma(y_i)-\frac{1}{n}\sum_{i=1}^n \ell_\sigma(y_i-f_\mathbf{z}(x_i))\leq  2 \sigma \|f_\mathbf{z}\|_\infty\leq 2\kappa\sigma\|f_\mathbf{z}\|_K. 
\end{align*}
As a result, we obtain
$\|f_\mathbf{z}\|_\infty\leq 2\kappa^2\sigma/\lambda$. This completes the proof of Lemma \ref{rough_bound}. 
\end{proof}

On the other hand, we can upper bound $f_{\sigma,\lambda}$ in the following way, which takes into account the heavy-tailedness of the distribution of the conditional variable $y|x$.
\begin{proposition}\label{upperbound_populationversion} 
Let $f_{\sigma,\lambda}$ defined in \eqref{population_version} be the population version of the $f_\mathbf{z}$.
We have
\begin{align*}
\|f_{\sigma,\lambda}\|_K\leq \sqrt{2 (\|a\|_{2,\rho}+1)}\sqrt{\frac{\sigma^{\max\{1-\epsilon,\,0\}} + \sigma^{1-\epsilon}}{\lambda}}.  
\end{align*}
\end{proposition}
\begin{proof}
According to the formulation in \eqref{population_version}, we know that 
\begin{align*}
\lambda\|f_{\sigma,\lambda}\|_K^2\leq \mathbb{E}\ell_\sigma(y).    
\end{align*}
We now bound $\mathbb{E}\ell_\sigma(y)$ by using a technique that is similar to the one developed in \cite{feng2020learning}. The idea is to decompose the product of the input and the output space into two regions, namely, $${\rm{I}}_{xy}=\left\{(x,y)||y|\leq \sigma\right\}$$ and $${\rm{II}}_{xy}=\left\{(x,y)||y|> \sigma\right\}.$$
As a result, we can represent and upper bound $\mathbb{E}\ell_\sigma(y)$ as follows:
\begin{align*}
\mathbb{E}\ell_\sigma(y)&=\int \int_{\mathrm{I}_{xy}} y^2 \mathrm{d} \rho(y|x)\mathrm{d}\rho_x(x) + \int \int_{\mathrm{II}_{xy}} (2\sigma|y|-\sigma^2) \mathrm{d} \rho(y|x)\mathrm{d}\rho_x(x)\\
&\leq \int\int_{\mathrm{I}_{xy}} y^2 \mathrm{d} \rho(y|x)\mathrm{d}\rho_x(x) + 2\sigma \int\int_{\mathrm{II}_{xy}} |y|  \mathrm{d} \rho(y|x)\mathrm{d}\rho_x(x).
\end{align*}
To proceed, we bound the two terms of the right-hand side of the above inequality, respectively. The first term can be upper bounded as follows when $0<\epsilon<1$:
\begin{align*}
\int\int_{\mathrm{I}_{xy}} y^2 \mathrm{d} \rho(y|x)\mathrm{d}\rho_x(x)&\leq \sigma^{1-\epsilon} \int\int_{\mathrm{I}_{xy}} |y|^{1+\epsilon} \mathrm{d} \rho(y|x)\mathrm{d}\rho_x(x)\\
&\leq \sigma^{1-\epsilon} \int\int_{|y|\leq \sigma} |y|^{1+\epsilon}\mathrm{d} \rho(y|x)\mathrm{d}\rho_x(x)\\
&\leq \sigma^{1-\epsilon} \int a(x) \mathrm{d} \rho_x(x)\\
&\leq \sigma^{1-\epsilon} \|a\|_{2,\rho},
\end{align*}
where $\|a\|_{2,\rho}=\sqrt{\int a(x)^2\mathrm{d}\rho_x(x)}$ and the last inequality is due to H\"older's inequality. When $\epsilon\geq 1$, the first term can be upper bounded by applying H\"older's inequality, which gives  
\begin{align*}
\int\int_{\mathrm{I}_{xy}} y^2 \mathrm{d} \rho(y|x)\mathrm{d}\rho_x(x)\leq \int (a(x))^{\frac{2}{1+\epsilon}}\mathrm{d}\rho_x(x)\leq \|a\|_{2,\rho}^{\frac{2}{1+\epsilon}}. 
\end{align*}
Combining the above two estimates for the case $\epsilon<1$ and the case $\epsilon\geq 1$ yields
\begin{align}\label{temp_initialbound_term1}
\int\int_{\mathrm{I}_{xy}} y^2 \mathrm{d} \rho(y|x)\mathrm{d}\rho_x(x) \leq \sigma^{\max\{1-\epsilon,\,0\}}(\|a\|_{2,\rho}+1).
\end{align}

We next bound the second term 
$$2\sigma \int\int_{\rm{II}_{xy}} |y|  \mathrm{d} \rho(y|x)\mathrm{d}\rho_x(x).$$  
By Assumption \ref{moment_assumption} and Markov's inequality  we have 
\begin{align*} 
\Pr\Big(|y|>\sigma |x\Big)\leq \cfrac{a(x)}{\sigma^{1+\epsilon}}.
\end{align*} 
By H\"older's inequality, we have
	\begin{align*} 
	    \int_{|y|>\sigma} |y|\mathrm{d}\rho(y|x) \leq \bigg(\Pr\Big(|y|>\sigma|x\Big) \bigg)^{\frac{\epsilon}{1+\epsilon}}
	    \bigg [\mathbb{E}\Big[|y|^{1+\epsilon}|x\Big]\bigg]^{\frac{1}{1+\epsilon}}
	    \leq \cfrac{a(x)}{\sigma^{\epsilon}}. 
\end{align*}
Consequently, we have 
\begin{align}\label{temp_initialbound_term2}
2\sigma \int\int_{\mathrm{II}_{xy}} |y|  \mathrm{d} \rho(y|x)\mathrm{d}\rho_x(x)=2\sigma\int\int_{|y|\geq \sigma}|y| \mathrm{d} \rho(y|x)\mathrm{d}\rho_x(x) \leq 2\sigma^{1-\epsilon} \|a\|_{2,\rho}.  
\end{align}
Combining the two estimates  in \eqref{temp_initialbound_term1} and \eqref{temp_initialbound_term2} yields 
\begin{align*}
\mathbb{E}\ell_\sigma(y)\leq 2 (\|a\|_{2,\rho}+1)(\sigma^{\max\{1-\epsilon,\,0\}} + \sigma^{1-\epsilon}).
\end{align*}
Therefore, we come to the following estimate
\begin{align*}
\lambda \|f_{\sigma,\lambda}\|_K^2\leq (\|a\|_{2,\rho}+1)(\sigma^{\max\{1-\epsilon,\,0\}} + \sigma^{1-\epsilon}),    
\end{align*}
which completes the proof of Proposition \ref{upperbound_populationversion}. 
\end{proof}

Note that in Lemma \ref{rough_bound}, we established that
\[
\|f_\mathbf{z}\|_\infty \leq \frac{\kappa^2 \sigma}{\lambda},
\]
while Proposition \ref{upperbound_populationversion} shows that
\[
\|f_{\sigma,\lambda}\|_K \leq \sqrt{2\big(\|a\|_{2,\rho}+1\big)} \sqrt{\frac{\sigma^{\max\{1-\epsilon,\,0\}} + \sigma^{1-\epsilon}}{\lambda}}.
\]
Since $\lambda$ must tend to zero to ensure the consistency of the regularization schemes \eqref{population_version} and \eqref{empirical_target_function}, and given that $\epsilon > 0$ and $\sigma > 1$, it is clear that the estimate for $f_{\sigma,\lambda}$ is considerably sharper than that for $f_{\mathbf{z}}$. One might expect roughly the same deterministic bound for both functions, as $f_{\sigma,\lambda}$ is simply the sample-free version of $f_{\mathbf{z}}$. However, due to the gap between $\mathbb{E}\ell_\sigma(y)$ and $\frac{1}{n}\sum_{i=1}^n \ell_\sigma(y_i)$, an analogous deterministic estimate for $f_{\mathbf{z}}$ is not available. Fortunately, $f_{\mathbf{z}}$ can be confined to the same hypothesis space in the following probabilistic manner:

\begin{theorem}\label{initial_upperbound}
Let $f_{\mathbf{z}}$ be produced by the regularized empirical risk minimization scheme \eqref{empirical_target_function}. Under Assumption \ref{moment_assumption}, for any $0<\delta<1$, with probability $1-\delta$, we have
\begin{align*}
\|f_{\mathbf{z}}\|_K\leq     \sqrt{2 (\|a\|_{2,\rho}+1)}\sqrt{\frac{\sigma^{\max\{1-\epsilon,\,0\}} + \sigma^{1-\epsilon}}{\delta\lambda}}.
\end{align*}
\end{theorem}
\begin{proof}
Recall the definition of $f_{\mathbf{z}}$, we have 
\begin{align*}
\lambda  \|f_{\mathbf{z}}\|_K^2 + \frac{1}{n}\sum_{i=1}^n \ell_\sigma(y_i-f_{\mathbf{z}}(x_i))\leq \frac{1}{n}\sum_{i=1}^n \ell_\sigma(y_i).    
\end{align*}
Since $(x_i,y_i)$, $i=1,\cdots, n$, are i.i.d., we have
\begin{align*}
\mathbb{E}\left(\frac{1}{n}\sum_{i=1}^n \ell_\sigma(y_i)\right)=\mathbb{E}\ell_\sigma(y).    
\end{align*}
From the proof of Proposition \ref{upperbound_populationversion}, we have
\begin{align*}
\mathbb{E}\ell_\sigma(y)\leq 2 (\|a\|_{2,\rho}+1)(\sigma^{\max\{1-\epsilon,\,0\}} + \sigma^{1-\epsilon}).
\end{align*}
By Markov's inequality, for any $0<\delta<1$, with probability at least $1-\delta$, we have
\begin{align*}
\frac{1}{n}\sum_{i=1}^n\ell_\sigma(y_i)\leq \frac{\mathbb{E}\ell_\sigma(y)}{\delta}\leq (2 (\|a\|_{2,\rho}+1))\frac{\sigma^{\max\{1-\epsilon,\,0\}} + \sigma^{1-\epsilon}}{\delta},
\end{align*}
which implies 
\begin{align*}
\lambda  \|f_{\mathbf{z}}\|_K^2 &\leq \lambda  \|f_{\mathbf{z}}\|_K^2 + \frac{1}{n}\sum_{i=1}^n \ell_\sigma(y_i-f_{\mathbf{z}}(x_i))\\
&\leq \frac{1}{n}\sum_{i=1}^n \ell_\sigma(y_i)\leq (2 (\|a\|_{2,\rho}+1))\frac{\sigma^{\max\{1-\epsilon,\,0\}} + \sigma^{1-\epsilon}}{\delta}.   
\end{align*}
We thus obtain the desired estimate
\begin{align*}
\|f_{\mathbf{z}}\|_K\leq     \sqrt{2 (\|a\|_{2,\rho}+1)}\sqrt{\frac{\sigma^{\max\{1-\epsilon,\,0\}} + \sigma^{1-\epsilon}}{\delta\lambda}}.
\end{align*}
This completes the proof of Theorem \ref{initial_upperbound}. 
\end{proof}

In fact, the probabilistic estimate in Theorem \ref{initial_upperbound} is crucial in the analysis as it inspires us to introduce a probabilistic effective hypothesis space to squeeze $f_\mathbf{z}$ into a more compact hypothesis space in a probabilistic way. 

\subsection{Conquering the unbounded and the heavy-tailed: The proposed strategies}
In this subsection, in an intuitive way, we briefly summarize our strategies for overcoming the challenges posed by an unbounded hypothesis space and heavy-tailed conditional distributions in robust regression learning. We achieve this by comparing our approach to the analysis of an unregularized robust regression learning scheme. The summary listed in this subsection will also serve as an outline of the subsequent analysis of this paper.

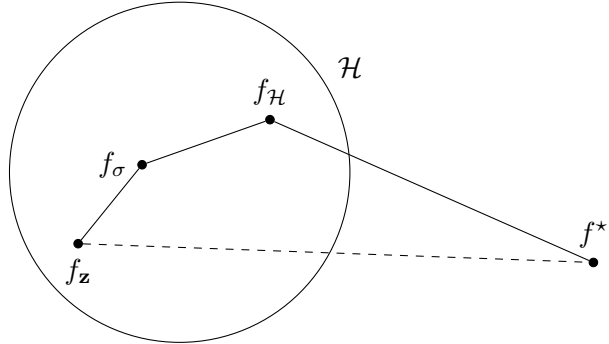
\begin{figure}[H]
\tikzset{trim left=0.0 cm}
\begin{tikzpicture}[xshift={0.5\textwidth-1.4cm}]
   \node [draw, circle, inner sep=1.6cm, label={30:$\mathcal{H}$}] (A) {};
   \node at (-1.35, -0.95) [draw, circle, inner sep=0.4mm, fill=black, label={-90:$f_{\mathbf{z}}$}] (D) {};
   \node at (1.2, 0.7) [draw, circle, inner sep=0.4mm, fill=black, label={90:$f_\mathcal{H}$}] (F) {};
     \node at (-0.5, 0.1) [draw, circle, inner sep=0.4mm, fill=black, label={180:$f_{\sigma}$}] (G) {};
   \node at (5.5, -1.2) [draw, circle, inner sep=0.4mm, fill=black, label={$f^\star$}] (E) {};
   \draw (F) -- (E)node[above,pos=0.45]{};
   \draw  (D) -- (G)node[above,pos=0.45]{};
   \draw  (G) -- (F)node[below,pos=0.50]{};
   \draw [dashed](D) -- (E);
\end{tikzpicture}
\caption{An illustration of the approach to bounding the prediction error $\|f_\mathbf{z}-f^\star\|_{2,\rho}^2$ when $f_\mathbf{z}$ is produced in \eqref{bounded_case} and the hypothesis space $\mathcal{H}$ is uniformly bounded. Here, $f_\sigma$ denotes the sample-free version of $f_\mathbf{z}$ and $f_{\mathcal{H}}$ denotes the projection of $f^\star$ onto $\mathcal{H}$.\label{analysis_illustration_bounded}}
\end{figure}

We begin with the following unregularized formulation of robust regression learning:
\begin{align}\label{bounded_case}
f_\mathbf{z} = \arg\min_{f\in\mathcal{H}} \frac{1}{n} \sum_{i=1}^n \ell_\sigma\big(y_i - f(x_i)\big),
\end{align}
where \(\mathcal{H}\) denotes a hypothesis space consisting of uniformly bounded functions. In this case, as shown in \cite{feng2022statistical1}, the analysis of $\|f_\mathbf{z}-f^\star\|_{2,\rho}^2$ can be carried out by bounding the distance between $f_\mathbf{z}$ and its sample-free version $f_\sigma$, the distance between $f_\sigma$ and $f_\mathcal{H}$, which is the projection of $f^\star$ onto $\mathcal{H}$, and the distance between $f_\mathcal{H}$ and $f^\star$, as illustrated in Figure \ref{analysis_illustration_bounded}. Specifically, we have
\begin{align}\label{bias-variance}
\|f_\mathbf{z}-f^\star\|_{2,\rho} \leq 
\underbrace{\|f_\mathbf{z}-f_\sigma\|_{2,\rho}}_{\text{(variance term)}}
\quad+\quad
\underbrace{\|f_\sigma-f_\mathcal{H}\|_{2,\rho}}_{\text{(bias term I)}}
\quad+\quad
\underbrace{\|f_\mathcal{H}-f^\star\|_{2,\rho}}_{\text{(bias term II)}},
\end{align}
where the variance term reflects the sample error caused by random samples and the choice of the hypothesis space $\mathcal{H}$, the bias term I is caused by the use of the robust loss function $\ell_\sigma$ and the integrated parameter $\sigma$, and the bias term II reflects the approximation ability of the tuple $(\rho,\mathcal{H}_K,\lambda)$ to $f^\star$. According to \cite{feng2022statistical1}, the estimation of the variance term and the bias term I relies on a comparison theorem that quantifies the difference between  $\mathcal{R}(f)-\mathcal{R}(f^\star)$ and $\|f-f^\star\|_{2,\rho}^2$ for  $f\in\mathcal{H}$. As pointed out above, the boundedness of functions in $\mathcal{H}$   plays an important role in establishing a useful upper bound of this difference.

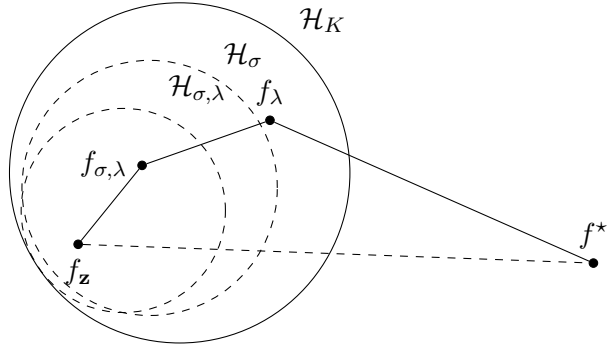
\begin{figure}[H]
\tikzset{trim left=0.0 cm}
\begin{tikzpicture}[xshift={0.5\textwidth-1.4cm}]
   % Largest hypothesis space: H_K
   \node [draw, circle, inner sep=1.6cm, label={50:$\mathcal{H}_K$}] (A) at (0,0) {};
   % New intermediate hypothesis space: H_\sigma, which includes the smallest circle and is contained in H_K
   \node [draw, dashed, circle, inner sep=1.2cm,  label={60:$\mathcal{H}_{\sigma}$}] (C) at (-0.4,-0.2) {};
   % Existing smallest hypothesis space: H_{\sigma,\lambda}, fully contained in H_\sigma
   \node [draw, dashed, circle, inner sep=0.96 cm, label={70:$\mathcal{H}_{\sigma,\lambda}$}] (B) at (-0.75,-0.5) {};
   % f_{\mathbf{z}}: placed inside H_{\sigma,\lambda}
   \node [draw, circle, inner sep=0.4mm, fill=black, label={-90:$f_{\mathbf{z}}$}] (D) at (-1.35, -0.95) {};
   % f_{\sigma,\lambda}
   \node [draw, circle, inner sep=0.4mm, fill=black, label={180:$f_{\sigma,\lambda}$}] (G) at (-0.5, 0.1) {};
   % f_\lambda
   \node [draw, circle, inner sep=0.4mm, fill=black, label={90:$f_\lambda$}] (F) at (1.2, 0.7) {};
   % f^\star
   \node [draw, circle, inner sep=0.4mm, fill=black, label={$f^\star$}] (E) at (5.5, -1.2) {};
   
   \draw (F) -- (E) node[above,pos=0.45]{};
   \draw (D) -- (G) node[above,pos=0.45]{};
   \draw (G) -- (F) node[below,pos=0.50]{};
   \draw [dashed] (D) -- (E);
\end{tikzpicture}
\caption{An illustration of our proposed approach to bounding the prediction error $\|f_\mathbf{z}-f^\star\|_{2,\rho}^2$, where $f_\mathbf{z}$ is produced in \eqref{empirical_target_function} and functions in the hypothesis space $\mathcal{H}_K$ are non-uniformly bounded. Here, $f_{\sigma,\lambda}$ defined in \eqref{population_version_huber_regularized} denotes the sample-free version of $f_\mathbf{z}$ and $f_{\lambda}$ is the reference function given in \eqref{population_version_ls_regularized}. \label{regularized_figure}}
\end{figure}

In contrast, when analyzing the regularized robust learning scheme \eqref{empirical_target_function}, while one can still work with the bias-variance decomposition \eqref{bias-variance}, the non-uniform boundedness of functions in $\mathcal{H}_K$ hinders the establishment of a meaningful bias estimation of the difference between $\mathcal{R}(f)-\mathcal{R}(f^\star)$ and $\|f-f^\star\|_{2,\rho}^2$. Inspired by the estimate in Theorem \ref{initial_upperbound}, we introduce a \textit{probabilistic effective hypothesis space} $\mathcal{H}_\sigma$ of \eqref{empirical_target_function} that is defined as follows:  
\begin{align}
\mathcal{H}_\sigma=\left\{f\in\mathcal{H}_K\,\Big|\,\|f\|_\infty\leq \frac{\sigma}{2}\right\}.    
\end{align}
This construction is based on two key observations. First, according to the discussion in \cite{feng2025tikhonov}, in order to derive a meaningful estimate of the difference between $\mathcal{R}(f)-\mathcal{R}(f^\star)$ and $\|f-f^\star\|_{2,\rho}^2$, $\|f_\mathbf{z}\|_\infty$ needs to be less than or equal to $\mathcal{O}(\sigma)$; Second, the estimate in Theorem \ref{initial_upperbound} implies that with high probability, we have 
\begin{align*}
 f_\mathbf{z}\in\mathcal{H}_{\sigma,\lambda} =\left\{ f\in\mathcal{H}_K\Big|\|f\|_\infty\leq \kappa \sqrt{2 (\|a\|_{2,\rho}+1)}\sqrt{\frac{\sigma^{\max\{1-\epsilon,\,0\}} + \sigma^{1-\epsilon}}{\delta\lambda}}\right\},
\end{align*}
which further implies that with properly chosen $\sigma$ and $\lambda$ values, it is possible to squeeze $f_\mathbf{z}$ into $\mathcal{H}_\sigma$ in a high probability sense. Therefore, in what follows, we shall work with the probabilistic effective hypothesis space $\mathcal{H}_\sigma$ in our analysis. The overall analysis procedure is summarized in Figure \ref{regularized_figure}, where the dashed circles indicate that $f_\mathbf{z}$ is contained in both spaces with high probability.

\section{Robust Learning for Regression is Asymptotically Mean Calibrated}
In this section, we study the mean calibration problem of the robust regression learning scheme \eqref{empirical_target_function}, which is concerned with the difference between  $\mathcal R(f) - \mathcal R(f^\star)$ and $\|f-f^\star\|_{2, \rho}^2$. As stated above, the estimate of the difference between the two quantities is crucial in bounding the variance term and the bias term I defined above. In particular, $f_\mathbf{z}$ is said to be \textit{asymptotic mean calibrated} if $\mathcal{R}(f_\mathbf{z})\to\mathcal{R}(f^\star)$ implies $f_\mathbf{z}\to f^\star$ as $n\to\infty$, where $f^\star$ denotes the conditional mean function. Based on our discussion above, we shall conduct our analysis in the probabilistic hypothesis space 
$\mathcal{H}_\sigma=\left\{f\in\mathcal{H}_K\,\Big|\,\|f\|_\infty\leq \frac{\sigma}{2}\right\}.$  Our main result is the following novel comparison theorem: 

\begin{theorem} \label{thm:comparison} 
    Let $\sigma >\max(2M, 1)$ and assume that Assumption \ref{moment_assumption} holds. For any function $f\in\mathcal{H}_\sigma$, there exists $C>0$ such that
    \begin{equation} \label{eq:comparison}
        (1-\alpha)\|f-f^\star\|_{2, \rho}^2 -\frac {C}{\alpha\sigma^{2\epsilon}} 
        \le \mathcal R(f) - \mathcal R(f^\star) \le 
        (1+\alpha)\|f-f^\star\|_{2, \rho}^2 +\frac {C}{\alpha \sigma^{2\epsilon}} 
    \end{equation}
    holds for all $\alpha>0$, where the constant $C$ is independent of $\sigma, \alpha$ or $f$.
\end{theorem}

\begin{proof} First notice that 
\begin{align} \label{eq:squarediff}
    (y-f(x))^2- (y-f^\star(x))^2 = (f^\star(x)-f(x))(2y-f^\star(x)-f(x))
\end{align} 
and the fact $f^\star(x) = \int_{\mathbb R} y \mathrm{d}\rho(y|x)$ implies 
\begin{align*}
  \|f-f^\star\|_{2, \rho}^2 = \int\int   (y-f(x))^2- (y-f^\star(x))^2  \mathrm{d}\rho(y|x) \mathrm{d}\rho_x(x).
\end{align*} 
Therefore, by defining 
\begin{equation*}
    \mathbf I_{xy} = \{(x, y): |y|>\sigma/2 \} \hbox{ and  } \mathbf {II}_{xy} 
    =\{(x, y): |y|\le \sigma/2 \}
\end{equation*} and noticing 
$\ell_\sigma(y-f(x)) = (y-f(x))^2$ and $\ell_\sigma(y-f^\star(x)) = (y-f^\star(x))^2$ on $\mathbf {II}_{xy}$ under the assumption $\sigma>2M>2|f^\star(x)|$ and $|f(x)|\le \sigma/2$,
we can write 
\begin{align*}
   &  \left| \mathcal R(f)- \mathcal R(f^\star) -\|f-f^\star\|_{2, \rho}^2 \right|  \\
    =  & \left |\int\int \left( \ell_\sigma(y-f(x)) - \ell_\sigma(y-f^\star(x))\right) - \left( (y-f(x))^2- (y-f^\star(x))^2 \right) \mathrm{d}\rho(y|x) \mathrm{d}\rho_x \right|  \\
   \le  &  \int\int_{\mathbf I_{xy}} \left| \ell_\sigma(y-f(x)) - \ell_\sigma(y-f^\star(x))\right| \mathrm{d}\rho(y|x) \mathrm{d}\rho_x (x) \\
 &  + \int\int_{\mathbf I_{xy}} \left| (y-f(x))^2- (y-f^\star(x))^2 \right| \mathrm{d}\rho(y|x) \mathrm{d}\rho_x(x)  \\
   := & Q_1 + Q_2.
\end{align*}
For $Q_1,$ by the Lipschitz property of $\ell_\sigma$ and the fact 
\begin{equation} \label{eq:prI}
   \int_{|y|>\sigma/2} \mathrm{d}\rho(y|x) = \Pr(|y|>\sigma/2|x) \le 
    \frac{\mathbb E[|y|^{1+\epsilon}|x]}{(\sigma/2)^{1+\epsilon}} \le \frac{2^{1+\epsilon} a(x)} {\sigma^{1+\epsilon}},
\end{equation}
we have 
\begin{align*}
    Q_1 & \le \sigma \int |f(x)-f^\star(x)| \int_{|y|>\sigma/2} \mathrm{d}\rho(y|x) \mathrm{d}\rho_x(x) \\
   & \le   {2^{1+\epsilon}}{\sigma^{-\epsilon}} 
    \int |f(x)-f^\star(x)| a(x) \mathrm{d}\rho_x(x) \\  
 & \le  {2^{1+\epsilon}}{\sigma^{-\epsilon}} 
    \|a\|_{2,\rho} \| f-f^\star\|_{2,\rho} .
\end{align*}
For $Q_2,$ using \eqref{eq:squarediff}, \eqref{eq:prI}, and 
\begin{align}
    \int_{|y|>\sigma/2} |y| \mathrm{d}\rho(y|x) 
    & \le \left( \int_{|y|>\sigma/2 } |y|^{1+\epsilon} \mathrm{d}\rho(y|x)\right)^{\frac 1{1+\epsilon}}
    \left(\int_{|y|>\sigma/2} \mathrm{d}\rho(y|x)\right)^{\frac{\epsilon}{1+\epsilon}}  \nonumber\\
   & \le (a(x))^{\frac 1 {1+\epsilon}} 
    \left( \frac {2^{1+\epsilon}a(x)}{\sigma^{1+\epsilon}}\right)^{\frac {\epsilon}{1+\epsilon}} 
    = 2^{\epsilon}\sigma^{-\epsilon}a(x),
    \label{eq:yI}
\end{align}
we have 
\begin{align*}
    Q_2 & \le \int |f(x)-f^\star(x)| \int_{|y|>\sigma/2} (2|y|+\sigma) \mathrm{d}\rho(y|x) \\
    & \le 2^{2+\epsilon}\sigma^{-\epsilon} \int |f(x)-f^\star(x)|a(x) \mathrm{d}\rho_x(x) \\
    & \le 2^{2+\epsilon}\sigma^{-\epsilon} 
    \|a\|_{2, \rho} \|f-f^\star\|_{2, \rho}
\end{align*}
Combining the estimation for $Q_1$ and $Q_2$ and with 
$C=(2^{\epsilon}+2^{1+\epsilon})^2\|a\|_{2,\rho}^2$ we obtain 
$$ \left| \mathcal R(f)- \mathcal R(f^\star) -\|f-f^\star\|_{2, \rho}^2 \right| 
\le 2\sqrt{C} \sigma^{-\epsilon}\|f-f^\star\|_{2,\rho} 
\le \frac{C}{\alpha \sigma^{2\epsilon}} + \alpha \|f-f^\star\|_{2,\rho}^2,
$$
which implies the desired conclusion.
\end{proof}

\section{Error Analysis with Heavy-Tailed Noise and Error Bounds}

\subsection{Error analysis with heavy-tailed noise}
Denote $$\Xi(x, y) = \ell_\sigma(y-f(x)) - \ell_\sigma(y-f^\star(x)).$$

\begin{theorem} \label{thm:varbound}
Under the assumptions of Theorem \ref{thm:comparison}, we have 
\begin{enumerate}
    \item [(i)] if $\epsilon\le 1$ then
    $$ \mathbb E[\Xi^2] \preccurlyeq \sigma^{1-\epsilon} (M+\|f\|_\infty) \left(\mathbb E[\Xi]+4C\sigma^{-2\epsilon} \right)^{\frac 12} + (M+\|f\|_\infty)^2 \left(\mathbb E[\Xi]+4C\sigma^{-2\epsilon} \right) $$
    \item [(ii)] if $\epsilon>1$, then
    $$ \mathbb E[\Xi^2] \preccurlyeq (M+\|f\|_\infty)^{\frac 2{1+\epsilon}}
  \left(\mathbb E[\Xi]+4C\sigma^{-2\epsilon} \right)^{\frac {\epsilon}{1+\epsilon}} + (M+\|f\|_\infty)^2 \left(\mathbb E[\Xi]+4C\sigma^{-2\epsilon} \right) 
    $$
\end{enumerate}
\end{theorem}

\begin{proof} With the definition of $\mathbf I_{xy}$ and $\mathbf {II}_{xy}$ in the proof of Theorem \ref{thm:comparison}, by the Lipschitz property of $\ell_\sigma$ and the facts $\ell_\sigma(y-f(x))=(y-f(x))^2$ and $\ell_\sigma(y-f^\star(x))=(y-f^\star(x))^2$ on $\mathbf {II}_{xy}$ under the assumptions, we are able to write
\begin{align*}
    & \mathbb E\left[ \left(\ell_\sigma(y-f(x)) - \ell_\sigma(y-f^\star(x))\right)^2\right] \\
    \le & \sigma^2 \int |f(x)-f^\star(x)|^2 \int_{|y|>\sigma/2} \mathrm{d}\rho(y|x)\mathrm{d}\rho_x(x) \\
    & + \int\int_{|y|\le \sigma/2} \left|(y-f(x))^2-(y-f^\star(x))^2\right|^2 \mathrm{d}\rho(y|x)\mathrm{d}\rho_x(x) \\
   := &V_1+V_2. 
   \end{align*}

If $\epsilon\le 1$,
by using \eqref{eq:prI} it is easy to verify that 
$$V_1 \preccurlyeq  \sigma^{1-\epsilon}(M+\|f\|_\infty)  \|a\|_{2,\rho}   \|f-f^\star\|_{2,\rho} . $$
For $V_2,$  we use \eqref{eq:yI} and obtain       
\begin{align*}
  V_2   
   & \preccurlyeq  \int |f(x)-f^\star(x)|^2 \int_{|y|<\sigma/2} 
   (2|y|+M+\|f\|_\infty)^2\mathrm{d}\rho(y|x)\rho_x(x) \\ 
 & \preccurlyeq  (M+\|f\|_\infty) \sigma^{1-\epsilon} 
   \int |f(x)-f^\star(x))|a(x) \mathrm{d}\rho(y|x) \mathrm{d}\rho_x (x)
 + (M+\|f\|_\infty)^2 \|f-f^\star\|_{2,\rho}^2 \\
& \preccurlyeq  \sigma^{1-\epsilon}(M+\|f\|_\infty)
 \|a\|_{2,\rho}  \|f-f^\star\|_{2,\rho} 
+ (M+\|f\|_\infty)^2 \|f-f^\star\|_{2,\rho}^2.  
\end{align*}
Combining the estimates for $V_1$ and $V_2$, and using Theorem \ref{thm:comparison}, the desired conclusion in part (i) is proved.

If $\epsilon>1,$ note that 
$$V_1 \le 4 \int |f(x)-f^\star(x)|^2 \int_{|y|<\sigma/2} 
   |y|^2 \mathrm{d}\rho(y|x)\rho_x (x) $$
   and 
   $$ V_2 \preccurlyeq \int |f(x)-f^\star(x)|^2 \int_{|y|<\sigma/2} 
   |y|^2 \mathrm{d}\rho(y|x)\rho_x(x) + (M+\|f\|_\infty)^2 \|f-f^\star\|_{2,\rho}^2.
   $$
   By 
$$\int |y|^2\mathrm{d}\rho(y|x) \le \left( \int |y|^{1+\epsilon} \mathrm{d}\rho(y|x)\right)^{\frac 2{1+\epsilon}} =(a(x))^{\frac 2 {1+\epsilon}},$$ we can prove 
\begin{align*}
  \int |f(x)-f^\star(x)|^2 \int
   |y|^2 \mathrm{d}\rho(y|x)\rho_x(x) 
   & \preccurlyeq (M+\|f\|_\infty)^{\frac 2{1+\epsilon}} 
   \int |f(x)-f^\star(x)|^{\frac{2\epsilon}{1+\epsilon}} (a(x))^{\frac 2{1+\epsilon}} \mathrm{d}\rho_x\\
 & \preccurlyeq   (M+\|f\|_\infty)^{\frac 2{1+\epsilon}}
  \|f-f^\star\|_{2,\rho}^{\frac {2\epsilon}{1+\epsilon}} \|a\|_{2,\rho}^{\frac 2 {1+\epsilon}},
\end{align*}
which leads to the desired conclusion in part (ii).
\end{proof}

To proceed, we need to introduce the following capacity assumption. For any $\eta>0$, let $\mathcal N(\mathcal{B}_R, \eta)$ denote the covering number of $\mathcal{B}_R$ by the balls of radius $\eta$ in $C(\mathcal X)$, that is, 
\begin{align*} 
\mathcal N(\mathcal{B}_R, \eta) = \min\left\{
k\in \mathbb N: \hbox{there exist } f_j\in \mathcal{B}_R, j=1,\ldots, k \hbox{ such that } \mathcal{B}_R\subset \bigcup_{j=1}^k B(f_j, \eta) \right \}, 
\end{align*} 
where $B(f_j, \eta)= \{f\in C(\mathcal X): \|f-f_j\|_\infty<\eta\}.$ 
Our capacity condition is stated as follows.

\begin{assumption}\label{complexity_assumption} 
There exist constants $q>0$ and $c_K>0$ such that for all $R\ge1$ and $\eta>0$,
\[
\log \mathcal N\!\left(\mathcal{B}_R, \eta; \|\cdot\|_\infty\right)
\;\le\; c_K \left(\frac{R}{\eta}\right)^q.
\]
\end{assumption}
This above assumption on covering numbers is typical in 
statistical learning theory; see e.g., \cite{cucker2007learning, steinwart2008support} and references therein.

\begin{proposition}
\label{prop:sample-error-bound}
Assume $1\le R\le \sigma/(2\kappa)$. For any $\delta>0$, with confidence at least $1-\delta$, 
$$ \left|\frac 1 n \sum_{i=1}^n \Xi(x_i, y_i) - \mathbb E[\Xi] \right|
\leq \frac 12 \mathbb E [\Xi] + 2C \sigma^{-2\epsilon} +\gamma_*$$ 
holds for all $f\in\mathcal B_R,$
where $$ \gamma_* \preccurlyeq  \begin{cases}
\dfrac{\sigma R }{n^{1/(q+1)}} \log\dfrac 2\delta, & \hbox{ if } \epsilon\le 1 \\[2em]
\left(\dfrac{\sigma R}{n^{1/(1+q)}} + \left(\dfrac{\sigma^q R^{q+2\zeta}} n \right)^{1/(1+q+\zeta)} \right) \log\dfrac 2\delta,
& \hbox{ if } \epsilon>1.
\end{cases}$$

\end{proposition}

\begin{proof}
By Bernstein's inequality, we know that 
\begin{align}
    & \Pr\left\{ \dfrac{\frac 1n \sum \Xi(x_i, y_i) -\mathbb E[\Xi]} {\sqrt{\mathbb E[\Xi] + 4C\sigma^{-2\epsilon}+\gamma} } 
    \ge \sqrt{\gamma}\right\} \nonumber \\
     \le & 2\exp\left\{\dfrac{-n \gamma(\mathbb E[\Xi] +4C\sigma^{-2\epsilon}+\gamma)} {2\mathbf{var}[\Xi] +\frac 23 \sigma(M+\|f\|_\infty) \sqrt{\gamma}
    {\sqrt{\mathbb E[\Xi] + 4C\sigma^{-2\epsilon}+\gamma}}   
    } \right\}. \label{eq:bern}
\end{align}
Theorem \ref{thm:comparison} with $\alpha=1$ implies that $\mathbb E[\Xi]+C\sigma^{-2\epsilon}>0.$ Therefore,
$$\sqrt{\gamma} \sqrt{\mathbb E[\Xi] + 4C\sigma^{-2\epsilon}+\gamma} \le \mathbb E[\Xi] + 4C\sigma^{-2\epsilon}+\gamma.$$ 
The assumption $f\in\mathcal B_R$ implies $\|f\|_\infty\le \kappa\|f\|_K\le \kappa R\le \sigma/2.$
By Theorem \ref{thm:varbound} (i) and the fact 
$$\sigma^{-\epsilon} = \frac 1 {\sqrt C} \sqrt {C\sigma^{-2\epsilon}} \le \frac 1 {\sqrt C} \sqrt{\mathbb E[\Xi] + 4C\sigma^{-2\epsilon}+\gamma},$$ we have 
$$\mathbf {var}[\Xi] \preccurlyeq \sigma R ( \mathbb E[\Xi] + 4C\sigma^{-2\epsilon}+\gamma). $$
Combining the above estimations, we prove that 
\begin{align*}
     \Pr\left\{ \dfrac{\frac 1n \sum \Xi(x_i, y_i) -\mathbb E[\Xi]} {\sqrt{\mathbb E[\Xi] + 4C\sigma^{-2\epsilon}+\gamma} } 
    \ge \sqrt{\gamma}\right\} 
    \le & 2\exp\left\{\dfrac{-n \gamma} {C_1 \sigma R}  \right\}.
\end{align*}
Then a standard argument leads to 
\begin{align*}
     \Pr\left\{ \sup_{f\in\mathcal B_R} \dfrac{\frac 1n \sum \Xi(x_i, y_i) -\mathbb E[\Xi]} {\sqrt{\mathbb E[\Xi] + 4C\sigma^{-2\epsilon}+\gamma} } 
    \ge 4\sqrt{\gamma}\right\} 
    \le & 2\exp\left\{c_K\left(\frac{\sigma R}{\gamma}\right)^q - \dfrac{n \gamma} {C_1 \sigma R}  \right\},
\end{align*}
which leads to the desired conclusion.

If $\epsilon>1,$ since none of the two terms in the variance bound in Theorem \ref{thm:varbound} (ii) dominates, we need to consider two different situations. For this purpose, let 
\begin{align*}
    \mathcal B_{R,0} & = \left\{f\in\mathcal B_R: (M+\|f\|_\infty)^{\frac 2{1+\epsilon}}
  \left(\mathbb E[\Xi]+4C\sigma^{-2\epsilon} \right)^{\frac {\epsilon}{1+\epsilon}} \le  (M+\|f\|_\infty)^2 \left(\mathbb E[\Xi]+4C\sigma^{-2\epsilon} \right)\right\} \\
  \mathcal B_{R, 1} & = \left\{f\in\mathcal B_R: (M+\|f\|_\infty)^{\frac 2{1+\epsilon}}
  \left(\mathbb E[\Xi]+4C\sigma^{-2\epsilon} \right)^{\frac {\epsilon}{1+\epsilon}} > (M+\|f\|_\infty)^2 \left(\mathbb E[\Xi]+4C\sigma^{-2\epsilon} \right) \right\}.
\end{align*}
For $f\in\mathcal B_{R,0},$ we use the inequality \eqref{eq:bern} and the estimation 
$$\hbox{var}[\Xi] \preccurlyeq  (M+\|f\|_\infty)^2 \left(\mathbb E[\Xi]+4C\sigma^{-2\epsilon} \right) \preccurlyeq 
\sigma R \left(\mathbb E[\Xi]+4C\sigma^{-2\epsilon}\right) $$
to obtain 
\begin{align*}
     \Pr\left\{ \dfrac{\frac 1n \sum \Xi(x_i, y_i) -\mathbb E[\Xi]} {\sqrt{\mathbb E[\Xi] + 4C\sigma^{-2\epsilon}+\gamma} } 
    \ge \sqrt{\gamma}\right\} 
    \le & 2\exp\left\{\dfrac{-n \gamma} {C_{2,0} \sigma R}  \right\}
\end{align*}
for some $C_{2,0}>0.$
A covering number argument gives the uniform concentration inequality
\begin{align*}
     \Pr\left\{ \sup_{f\in\mathcal B_{R,0}} \dfrac{\frac 1n \sum \Xi(x_i, y_i) -\mathbb E[\Xi]} {\sqrt{\mathbb E[\Xi] + 4C\sigma^{-2\epsilon}+\gamma} } 
    \ge 4\sqrt{\gamma}\right\} 
    \le & 2\exp\left\{c_K\left(\frac{\sigma R}{\gamma}\right)^q - \dfrac{n \gamma} {C_{2,0} \sigma(M+\|f\|_\infty)}  \right\}
\end{align*}
which proves that the desired conclusion holds for all $f\in\mathcal B_{R, 0}.$

For $f\in\mathcal B_{R,1},$ denote $\zeta = \frac 1 {1+\epsilon}.$ 
We first use Bernstein inequality to obtain 
\begin{align}
    & \Pr\left\{ \dfrac{\frac 1n \sum \Xi(x_i, y_i) -\mathbb E[\Xi]} {\left({\mathbb E[\Xi] + 4C\sigma^{-2\epsilon}+\gamma}\right)^{{(1-\zeta)}/{2}} } 
    \ge \gamma^{(1+\zeta)/2} \right\} \nonumber \\
    \le & 2\exp\left\{\dfrac{-n \gamma ^{1+\zeta} (\mathbb E[\Xi] +4C\sigma^{-2\epsilon}+\gamma)^{1-\zeta}}  {2\mathbf{var}[\Xi] +\frac 23 \sigma(M+\|f\|_\infty) \gamma^{(1+\zeta)/2}
    (\mathbb E[\Xi] + 4C\sigma^{-2\epsilon}+\gamma)^{(1-\zeta)/2} 
    } \right\}. \label{eq:bern2}
\end{align}
Since $\zeta\le 1,$ it holds that
$$  \gamma^{(1+\zeta)/2}
    (\mathbb E[\Xi] + 4C\sigma^{-2\epsilon}+\gamma)^{(1-\zeta)/2} \le \gamma^{\zeta} (\mathbb E[\Xi] + 4C\sigma^{-2\epsilon}+\gamma)^{(1-\zeta)}. $$
By Theorem \ref{thm:varbound} (ii) and the definition of $\mathcal B_{R,1},$ we have 
\begin{align*}
    \hbox{var}[\Xi] & \preccurlyeq R^{2\zeta}
  \left(\mathbb E[\Xi]+4C\sigma^{-2\epsilon} + \gamma \right)^{1-\zeta}
\end{align*}
Therefore, there exists a constant $C_{2,1}>0$ such that
\begin{align*}
    & \Pr\left\{ \dfrac{\frac 1n \sum \Xi(x_i, y_i) -\mathbb E[\Xi]} {\left({\mathbb E[\Xi] + 4C\sigma^{-2\epsilon}+\gamma}\right)^{{(1-\zeta)}/{2}} } 
    \ge \gamma^{ (1+\zeta)/2} \right\} 
    \le  2\exp\left\{\dfrac{-n \gamma ^{1+\zeta} }  {C_{2, 1}(R^{2\zeta} + \sigma R \gamma^\zeta}) \right\}. 
\end{align*}
Again we use a covering number argument to obtain
\begin{align*}
    & \Pr\left\{ \sup_{f\in\mathcal B_{R,1}} \dfrac{\frac 1n \sum \Xi(x_i, y_i) -\mathbb E[\Xi]} {\left({\mathbb E[\Xi] + 4C\sigma^{-2\epsilon}+\gamma}\right)^{{(1-\zeta)}/{2}} } 
    \ge \gamma^{(1+\zeta)/2} \right\} 
    \le  2\exp\left\{c_K\left(\frac  {\sigma R}{\gamma} \right)^q - \dfrac{n \gamma ^{1+\zeta} }  {C_{2, 1}(R^{2\zeta} + \sigma R \gamma^\zeta)} \right\}. 
\end{align*}
It implies that the conclusion holds with probability $1-\delta$ with 
\begin{align*}
\gamma _* \preccurlyeq \left(\dfrac{\sigma R}{n^{1/(1+q)}} + \left(\dfrac{\sigma^q R^{q+2\zeta}} n \right)^{1/(1+q+\zeta)} \right) \log\frac 2\delta.
\end{align*}
Combining the estimates for both cases, we prove that the desired conclusion holds for all $f\in\mathcal B_R.$
\end{proof}

\subsection{Results on error bounds}
To derive the convergence rates, we need an assumption on the approximation ability 
of the reproducing Hilbert space. In this paper we adopt the following one in terms of the decay of a 
$K$-functional \cite{cucker2007learning}:
$$ \min_{f\in\mathcal H_K} \|f-f^\star\|_{2, \rho}^2 +\lambda \|f\|_K^2  \preccurlyeq \lambda ^\beta $$
for some $0<\beta \le 1.$ It holds when $f^\star$ lies in the interpolation space 
between $L^2$ and $\mathcal H_K$, which is usually measured by a fractional power of the 
kernel integral operator and stated as the so-called source conditions. 

Our results on error bounds are stated in the theorem below. 

\begin{theorem}
\label{thm:rate}
Let $\alpha=\frac{2\epsilon}{\beta} \wedge (1+\epsilon) \wedge 2$
and choose $\sigma=n^{\frac{2} {(1+q)(2+\alpha+\alpha\beta)}}$
    and $\lambda = \eta\sigma^{-\alpha}$ with 
    $\eta$ some absolute constant independent of $n, \lambda, \sigma$. Then, for any $0<\delta<1$, with probability at least $1-\delta$, it holds that 
    \begin{align*}
     \|f_{\mathbf{z}} -f^\star\|_{2,\rho}^2 
     \preccurlyeq \left(\log(8/\delta)+\log\log\log n\right)^2 \, n^{-\frac{2\alpha\beta} {(1+q)(2+\alpha+\alpha\beta)}} \log n .
    \end{align*}
\end{theorem}

The proof of Theorem \ref{thm:rate} needs the proposition below. 
\begin{proposition}
    \label{prop:reduceR}
Under the assumptions of Theorem \ref{thm:rate}, if $\|f_{\mathbf{z}}\|_K \le b\sigma^r \wedge  \frac{\sigma} {18\kappa}$ with probability $1-\delta_1$ for some $b>0$ and $1-\beta<r\le 1$, then for any $\tilde\delta>0$, with probability $1-\delta_1-\tilde \delta$, it holds that 
$$\|f_{\mathbf{z}}\|_K\le \sqrt{\tilde C (b\vee 1)\log(2/\tilde \delta)}\   \sigma^{\frac{r+\alpha(1-\beta)/2}{2}}.$$ 
\end{proposition}

\begin{proof} With the choice of the parameters $\sigma$ and $\lambda,$ and by the assumption $R\le b\sigma^r$, 
it is easy to verify that for all $\epsilon>0$ there hold
$\sigma^{-2\epsilon} \le \sigma ^{-\alpha \beta} $  and   $\gamma_* \preccurlyeq b \sigma^{r-\alpha(1+\beta)/2}.$
If $\alpha(1-\beta)/2 \le r \le 1,$ Proposition \ref{prop:sample-error-bound} implies that, there exists an absolute constant $C'>0$ such that for all $f\in \mathcal B_R,$ 
\begin{align} \label{eq:sample-error-bound}
    |(\mathcal R(f) -\mathcal R(f^\star)) -(\mathcal R_{\mathbf z}(f) - \mathcal R_{\mathbf z} (f^\star)) |
    \le \frac 12 (\mathcal R(f) -\mathcal R(f^\star)) + C' (b\vee 1) \log(\tfrac 2 \delta) \sigma^{r-\alpha(1+\beta)/2} 
\end{align}
holds with confidence at least $1-\delta.$

Let 
$$f_\lambda = \arg\min_{f\in \mathcal H_K} \left\{ \|f-f^\star\|_{2,\rho}^2 + \lambda \|f\|_K^2\right\}.$$ 
By the definition of $f_{\mathbf z},$ it holds that 
$$ \mathcal R_{\mathbf z} (f_{\mathbf z}) +\lambda \|f_{\mathbf z}\|_K^2 
\le \mathcal R_{\mathbf z} (f_{\lambda}) +\lambda \|f_{\lambda}\|_K^2 .$$
A standard error decomposition process (see e.g., \cite{wu2006learning, feng2022statistical1}) implies that 
\begin{align*} 
\mathcal R (f_{\mathbf z}) - \mathcal R (f^\star) +\lambda \|f_{\mathbf z}\|_K^2  
& \le 
\Big\{ \left(\mathcal R (f_{\mathbf z}) -\mathcal R(f^\star) \right) - 
\left(\mathcal R_{\mathbf z}(f_{\mathbf z}) - \mathcal R_{\mathbf z}(f^\star) \right) 
\Big\}  \\
&\qquad  + \Big\{ \left(\mathcal R_{\mathbf z} (f_{\lambda}) -\mathcal R_{\mathbf z} (f^\star) \right) - 
\left(\mathcal R(f_{\lambda}) - \mathcal R(f^\star) \right) 
\Big\}   \\
&\qquad + \mathcal R(f_\lambda) - \mathcal R(f^\star)+\lambda \|f_\lambda\|_K^2.
\end{align*}
Note that $\|f_\lambda\|_K \preccurlyeq \sigma^{\alpha(1-\beta)/2}$ and by \eqref{eq:comparison}
$$ \mathcal R(f_\lambda) - \mathcal R(f^\star)+\lambda \|f_\lambda\|_K^2 
\le 2 \|f_\lambda -f^\star\|_{2, \rho}^2 + \lambda \|f_\lambda\|_K^2 + C\sigma^{-2\epsilon}
\preccurlyeq \sigma^{-\alpha\beta}.$$

We can apply \eqref{eq:sample-error-bound} to $f_{\mathbf z}$ and $f_\lambda$ and obtain that,
with confidence at least $1-\delta_1 - \delta$
\begin{align*}
    \mathcal R (f_{\mathbf z}) - \mathcal R (f^\star) +\lambda \|f_{\mathbf z}\|_K^2  
    & \le \frac 12 \Big( \mathcal R (f_{\mathbf z} -\mathcal R(f^\star) \Big)
     +  C' (b\vee 1) \log(\tfrac 2 \delta) \sigma^{r-\alpha(1+\beta)/2},
\end{align*}
or equivalently, 
\begin{align} \label{eq:bound-in-sigma}
    \mathcal R (f_{\mathbf z}) - \mathcal R (f^\star) + 2\lambda \|f_{\mathbf z}\|_K^2  
    & \le  2 C'' (b\vee 1) \log(\tfrac 2 \delta) \sigma^{r-\alpha(1+\beta)/2}
\end{align}
for some absolute constant $C''>0.$ 
Applying \eqref{eq:comparison} again, we have 
$$ \frac 12 \|f_{\mathbf z}-f^\star\|_{2, \rho}^2 + 2\lambda \|f_{\mathbf z}\|_K^2  
 \le  2 C'' (b\vee 1) \log(\tfrac 2 \delta) \sigma^{r-\alpha(1+\beta)/2} + 2C \sigma^{-2\epsilon} $$
    holds with confidence at least $1-\delta_1 - \tilde \delta$, 
implying the desired conclusion with $\tilde C = C''+C.$
\end{proof}

It is important to notice that if $b\ge (\tilde C \vee 1)\log(\frac 2 {\tilde \delta}),$ then $\sqrt{\tilde C (b\vee 1)\log(2/\tilde \delta)} \le b$. Therefore, Proposition \ref{prop:reduceR} refines the bound of 
$\|f_{\mathbf z}\|_K$ by reducing the power index and maintaining the constant simultaneously 
at a price of sacrificing the confidence. We can apply the proposition repeatedly until the refinement stops as $r$ approaches $\frac {\alpha(1-\beta)}{2}.$

\begin{proof}[Proof of Theorem \ref{thm:rate}] Denote $r^*=\frac{\alpha(1-\beta)}{2}.$ We first apply Proposition \ref{initial_upperbound} and obtain that $$\|f_{\mathbf z}\|_K \le \frac {\sigma}{2\kappa}$$ with probability $1-\frac \delta 2.$ Then we apply Proposition \ref{prop:reduceR} iteratively and obtain that, for any $\tilde\delta >0$ and any integer $j>0,$ 
$$\|f_{\mathbf z}\|_K \preccurlyeq \log(\tfrac 2 {\tilde\delta}) \sigma ^{r^*+2^{-j}(1-r^*)}$$
with confidence at least $1-\frac \delta 2 -j \tilde \delta.$ 
Let $$j_0 = \left\lceil \log_2 \left(\frac{(1-r^*)\log\sigma}{\log\log\sigma}\right)\right\rceil. $$  
We have
$$\|f_{\mathbf z}\|_K \preccurlyeq \log(\tfrac 2 {\tilde\delta}) \sigma ^{r^*}\log \sigma$$
with confidence at least $1-\frac \delta 2 -j_0 \tilde \delta.$ 
Applying \eqref{eq:bound-in-sigma}, we have 
$$\mathcal R (f_{\mathbf z}) - \mathcal R (f^\star) \preccurlyeq \log(\tfrac 2 {\tilde\delta}) 
\sigma ^{-\alpha\beta}\log \sigma$$
with confidence at least $1-\frac \delta 2 -(j_0+1) \tilde \delta.$ 
Choosing $\tilde\delta=\frac{\delta}{2(j_0+1)}$ yields 
$$\|f_{\mathbf z}\|_K \preccurlyeq \log(\tfrac {4(j_0+1)} {\delta}) \sigma ^{r^*}\log \sigma 
\preccurlyeq \left(\log\tfrac{4}{\delta} +\log\log n\right) n^{-\frac{2\beta}{(1+q)(2+\alpha+\alpha\beta)}} \log n$$
with confidence at least $1-\delta.$ 

\end{proof}

\subsection{From the old to the new: Technical contributions}
Classical analyses of regression learning typically hinge on uniformly bounded hypothesis spaces, light-tailed noise, and excess-risk bounds as a proxy for prediction accuracy \cite{cucker2007learning, steinwart2008support}. In heavy-tailed, nonparametric regimes, these pillars no longer suffice. Our contribution is a new pipeline that replaces uniform boundedness by probabilistic localization, light-tailed concentration by weak-moment deviation bounds, and excess-risk surrogates by direct control of prediction error via explicit comparison inequalities, which, based on the theoretical results developed above, are detailed below.

\paragraph{(A) Localization via a probabilistic effective hypothesis space.}
We introduce the \emph{probabilistic effective hypothesis space}
\[
\mathcal{H}_\sigma \;=\; \Big\{ f\in\mathcal{H}_K : \|f\|_\infty \le \sigma/2 \Big\},
\]
and show that both the population solution $f_{\sigma,\lambda}$ and the empirical solution $f_{\mathbf z}$ enter a common, data–dependent sup‑norm region with high probability. Concretely, Proposition~\ref{upperbound_populationversion} and Theorem~\ref{initial_upperbound} yield that with probability at least $1-\delta$, it holds that 
\[
\|f_{\sigma,\lambda}\|_K \ \vee \ \|f_{\mathbf z}\|_K
\;\;\preccurlyeq \;\;
\sqrt{\frac{\sigma^{\max\{1-\epsilon,\,0\}} + \sigma^{1-\epsilon}}{\delta\,\lambda}},
\]
and hence, by the RKHS embedding $\|f\|_\infty\le \kappa\|f\|_K$, 
$f_{\mathbf z}\in\mathcal{H}_\sigma$ holds with probability at least $1-\delta$ whenever $\lambda$ scales with $\sigma$ so that
\[
\kappa\sqrt{\frac{\sigma^{\max\{1-\epsilon,\,0\}} + \sigma^{1-\epsilon}}{\delta\,\lambda}}
\;\le\; \sigma/2.
\]
For instance, taking $\lambda=\eta\sigma^{-\alpha}$ with $\alpha\le 1+\epsilon$ when $\epsilon\in(0,1)$ and $\alpha\le 2$ when $\epsilon\ge 1$ suffices. This localization is the backbone that restores concentration without any \emph{a priori} uniform boundedness of the hypothesis space.

\paragraph{(B) Uniform comparison between robust risk and prediction error.}
On $\mathcal{H}_\sigma$ we prove a two–sided comparison (Theorem~\ref{thm:comparison}): for any $\alpha>0$ and any $f\in\mathcal{H}_\sigma$,
\[
(1-\alpha)\|f-f^\star\|_{2,\rho}^2 \;-\; \frac{C}{\alpha\,\sigma^{2\epsilon}}
\ \le\ 
\mathcal R(f)-\mathcal R(f^\star)
\ \le\ 
(1+\alpha)\|f-f^\star\|_{2,\rho}^2 \;+\; \frac{C}{\alpha\,\sigma^{2\epsilon}}.
\]
This establishes \emph{asymptotic mean calibration} within $\mathcal{H}_\sigma$: reductions in excess robust risk translate into reductions in $L_2$ prediction error up to an explicit residual of order $\sigma^{-2\epsilon}$. The residual quantifies the robustness–bias induced by the scale parameter and vanishes as $\sigma\to\infty$ at a rate determined by the tail parameter $\epsilon$.

\paragraph{(C) Heavy‑tail–aware variance control and uniform deviations.}
Let $\Xi(x,y)=\ell_\sigma(y-f(x))-\ell_\sigma(y-f^\star(x))$. We derive second‑moment bounds that adapt to the tail regime (Theorem~\ref{thm:varbound}):
\[
\mathbb E[\Xi^2]
\;\preccurlyeq \;
\begin{cases}
\sigma^{1-\epsilon}(M+\|f\|_\infty)\,\big(\mathbb E[\Xi]+4C\sigma^{-2\epsilon}\big)^{1/2}
+(M+\|f\|_\infty)^2\big(\mathbb E[\Xi]+4C\sigma^{-2\epsilon}\big), & \epsilon\le 1,\\[0.4em]
(M+\|f\|_\infty)^{\frac{2}{1+\epsilon}}\big(\mathbb E[\Xi]+4C\sigma^{-2\epsilon}\big)^{\frac{\epsilon}{1+\epsilon}}
+(M+\|f\|_\infty)^2\big(\mathbb E[\Xi]+4C\sigma^{-2\epsilon}\big), & \epsilon>1.
\end{cases}
\]
Combined with a covering number condition for RKHS balls (with exponent $q$), this yields a \emph{uniform} Bernstein‑type deviation bound (Proposition~\ref{prop:sample-error-bound}): for any radius $R\le \sigma/(2\kappa)$,
\[
\sup_{f\in\mathcal B_R}
\left|\frac1n\sum_{i=1}^n \Xi(x_i,y_i) - \mathbb E[\Xi]\right|
\ \le\ 
\frac12\,\mathbb E[\Xi] \;+\; 2C\sigma^{-2\epsilon} \;+\; \gamma_\ast,
\]
with an explicit $\gamma_\ast$ that depends on $(\sigma,R,n,q)$ and the tail regime. This step furnishes the variance term in our bias–variance analysis \emph{without} requiring finite second moments.

Taken together, (A)–(C) yield a unified pipeline for robust nonparametric regression under weak $(1+\epsilon)$‑moment assumptions: high‑probability localization to $\mathcal{H}_\sigma$, uniform comparison of robust risk and prediction error with an explicit $\sigma$‑controlled residual, and heavy‑tail–aware uniform deviations on localized classes. Beyond the specific Huber–RKHS instance, these ingredients are modular and readily transferable: the localization step only requires a high‑probability sup‑norm control; the comparison step hinges on linear‑growth tails and local quadratic behavior of the loss, covering widely used convex and nonconvex robust losses; and the deviation step couples weak‑moment tail bounds with standard complexity measures and can be adapted to other function classes. Consequently, the same proof skeleton extends to a broad family of robust risk minimization schemes and yields explicit, finite‑sample guarantees under weak moments. Practically, it provides principled joint tuning rules for $(\sigma,\lambda)$ and a clear diagnostic for learnability in terms of prediction error rather than excess robust risk, which, together, thereby deliver a general methodology for understanding robust machine learning for nonparametric regression with heavy‑tailed noise.

\section{Extended Discussion and Perspectives}
In this section, we broaden the scope of our analysis to situate our findings within the larger context of robust machine learning for nonparametric regression. 

\subsection{How does nonparametric Huber regression outclass least squares approaches?}

When errors are light‑tailed, e.g., sub‑Gaussian or sub‑exponential with finite variance, nonparametric Huber regression can be tuned so that it behaves essentially like least squares, while retaining protection against occasional anomalies. In our study, taking the robustness scale $\sigma$ large and coupling it with $\lambda$ ensures that the loss is quadratic on the bulk of the residuals and the robust bias is negligible. The comparison result (Theorem~\ref{thm:comparison}) shows that excess Huber risk and $L_2$ prediction error coincide up to a vanishing $\sigma^{-2\epsilon}$ term, and the rate theorem (Theorem~\ref{thm:rate}) recovers the same oracle‐type convergence rates (up to log factors) that kernel ridge regression attains under standard capacity and approximation conditions. Thus, in light‑tailed regimes, Huber regression matches least squares in rate while offering additional stability to small departures from ideality such as contamination, mild outliers, or model misspecification.

The contrast is sharper under heavy‑tailed noise. If only a weak $(1+\epsilon)$‑moment exists, the squared‑loss risk may be undefined or fail to concentrate, and empirical least squares can become statistically and computationally unstable. In precisely this setting, Huber regression remains well‑posed and mean‑calibrated on a localized hypothesis space: Theorem~\ref{thm:comparison} links robust excess risk to prediction error with an explicit, tail‑dependent residual, and our heavy‑tail deviation bounds are established without requiring finite second moments. The resulting nonasymptotic guarantees in Theorem~\ref{thm:rate} show that, with a diverging $\sigma$ and appropriately decaying $\lambda$, Huber regression achieves fast prediction rates that are unattainable by unmodified least squares in the same conditions. In short, when errors are light‑tailed, Huber is competitive with least squares; when errors are heavy‑tailed, least squares may be incapable or require ad‑hoc fixes, whereas Huber continues to deliver stable estimation with provable rates.

\subsection{Robustification in robust learning: Trading bias for robustness}

Robust regression schemes, exemplified here by nonparametric Huber regression, mitigate the influence of outliers and heavy-tailed noise by tempering the contribution of large residuals. The price of this robustification is a systematic bias relative to the least‑squares target. Thus, robustification is inherently a trade: variance is reduced through bounded influence on extreme errors, while a robust bias is introduced.

For the Huber loss $\ell_\sigma$, small residuals incur a quadratic penalty (as in least squares) whereas large residuals are penalized linearly. This ``quadratic–linear’’ transition at scale $\sigma$ not only caps the effective influence of heavy tails, but also shifts the objective away from the pure $L_2$ criterion. Our theory makes this trade‑off explicit. On the localized class $\mathcal{H}_\sigma$, Theorem~\ref{thm:comparison} shows that the excess robust risk and the $L_2$ prediction error are equivalent up to a loss‑induced residual of order $\sigma^{-2\epsilon}$. In particular, decreasing $\sigma$ strengthens robustness but enlarges this residual bias; increasing $\sigma$ weakens robustness but shrinks the residual, recovering the squared‑loss target in the limit. Critically, $\sigma$ also governs the size of the effective hypothesis space: larger $\sigma$ enlarges $\mathcal{H}_\sigma=\{f:\|f\|_\infty\le \sigma/2\}$ and can magnify sample error, while smaller $\sigma$ tightens $\mathcal{H}_\sigma$ and aids concentration. This creates a \emph{two‑lever} design problem, $\sigma$ for robustness and $\lambda$ for regularization, that must be tuned jointly. Our high‑probability localization bounds (Proposition~\ref{upperbound_populationversion}, Theorem~\ref{initial_upperbound}) and tail‑aware deviation inequalities (Proposition~\ref{prop:sample-error-bound}) quantify the variance component on $\mathcal{H}_\sigma$, while the comparison theorem quantifies the robustness bias via the explicit $\sigma^{-2\epsilon}$ term. The rate result (Theorem~\ref{thm:rate}) then prescribes a coordinated choice in which $\sigma$ increases with $n$ (so the robustness bias vanishes) and $\lambda$ decreases accordingly (so approximation and variance remain balanced), yielding nonasymptotic prediction guarantees under weak $(1+\epsilon)$‑moment conditions. In this sense, $\sigma$ plays a role akin to the regularization parameter $\lambda$. While $\lambda$ penalizes the RKHS norm and shrinks the estimator toward smoother functions, $\sigma$ regularizes along two complementary axes: it shapes the loss's influence of large residuals and shrinks the effective search region through the sup‑norm constraint $\|f\|_\infty\!\le\!\sigma/2$ that defines $\mathcal{H}_\sigma$. Thus the learned empirical target function lies in a doubly constrained set controlled by $\sigma$ and $\lambda$, revealing that they both act as capacity controls. Each induces bias: the \emph{Tikhonov bias} decays as $\lambda^\beta$, while the \emph{robustness bias} decays as $\sigma^{-2\epsilon}$ (see Theorem~\ref{thm:comparison}), both of which affect variance.  

Practically, our analysis shows how tail heaviness, hypothesis‑class complexity, and approximation smoothness jointly determine the optimal bias–variance balance. Robustification should be understood as introducing a controlled, ultimately vanishing bias that buys a substantial variance reduction and restores concentration, thereby improving prediction error over least squares in heavy‑tailed, nonparametric settings. Moreover, $\sigma$ functions as a robustness regularizer complementary to $\lambda$: decreasing $\sigma$ plays a role akin to increasing $\lambda$  whereas increasing $\sigma$ mirrors decreasing $\lambda$. In short, robustification explicitly \emph{trades bias for robustness}; by calibrating $\sigma$ together with $\lambda$, this trade becomes quantifiable and favorable in the heavy‑tailed regimes that motivate robust learning.

\subsection{Insights from nonparametric Huber regression into general robust learning} 

Our analysis of nonparametric Huber regression yields several lessons that extend to robust learning more broadly. 

First, allowing the robustness scale to diverge is essential for reconciling robustness in conditional mean-based prediction. In Huber’s loss, the scale controls where the penalty transitions from quadratic to linear. If the scale is kept fixed, the objective does not fully recover the quadratic behavior on moderate residuals, and a non‑negligible robust bias remains. Letting the scale grow with the sample size drives that bias to zero while retaining protection against rare, extreme errors, an effect that is specific to modern nonparametric regimes where function‑class complexity is high and differs from classical parametric M‑estimation, where $n\!\gg\!p$ and a fixed scale is often adequate.

Second, as mentioned above, robustification is also a form of capacity control. By tuning the scale parameter, we indirectly regulate the effective hypothesis space through a high‑probability sup‑norm localization. Larger scales relax this localization and can increase variance; smaller scales tighten it and promote concentration. Thus, robustness tuning and regularization interact to jointly balance approximation error and sample error. Our study makes this joint control explicit and provides concrete coupling rules for the scale and the regularization strength.

Third, the local behavior of the robust loss near zero is pivotal: to target mean‑squared prediction error and obtain fast rates, the loss should be  quadratic in a neighborhood of zero. This local quadraticity guarantees convexity within the localized class and ensures mean‑calibration: small residuals are penalized like in least squares, so the population minimizer tracks the conditional mean when the robust bias vanishes and leads to fast convergence rates under weak moments. 

\section{Conclusion}

This paper develops a unified theoretical framework for robust nonparametric regression in the presence of heavy-tailed noise and unbounded hypothesis spaces. Anchored in Huber regression within an RKHS, our analysis tackles two core obstacles: the failure of light-tailed concentration and the lack of uniform boundedness. The key device is a \emph{probabilistic effective hypothesis space} that localizes the learned predictor with high probability, making meaningful concentration and complexity control possible without global sup‑norm assumptions. On this localized class we establish comparison results that link excess robust risk to $L_2$ prediction error up to an explicit, vanishing residual, thereby identifying prediction error—not excess robust risk—as the correct target of analysis in robust learning. Building on these ingredients, we derive nonasymptotic error bounds and convergence rates under weak $(1+\epsilon)$-moment conditions, clarifying how tail heaviness, function-class complexity, and approximation smoothness together determine performance. Taken together, these results formalize when and why robust regression improves upon least squares in nonparametric, heavy‑tailed regimes, and they provide principled guidance for jointly tuning robustness and regularization. Beyond RKHSs, the same strategy should transfer to other nonparametric models with suitable capacity measures, including overparameterized networks, where implicit or explicit mechanisms such as early stopping, weight decay, and gradient clipping  can play the role of our effective localization. We view the framework here as a step toward a general theory of robust learning under weak moments: modular technical tools, calibrated performance metrics, and explicit trade‑offs that together translate robustness from a loss‑level heuristic into a principled, rate‑optimal methodology for machine learning for regression.

\section*{Acknowledgments}
The work of Yunlong Feng is partially supported by NSF DMS-2111080. The work of Qiang Wu is partially supported by NSF DMS-2448749. The two authors made equal contributions to this paper and are listed alphabetically.

\bibliographystyle{plain}
\bibliography{FENGWU25}

\end{document}